\documentclass[twoside]{article}

\usepackage[accepted]{style}
\usepackage[round]{natbib}

\usepackage{amsmath,amssymb}
\usepackage{graphicx} 
\usepackage{hyperref}
\usepackage{bbm}
\usepackage{siunitx}
\usepackage{subcaption}
\usepackage{pgffor}
\usepackage{tcolorbox}
\usepackage{tikz}
\usepackage{pgfplots}
\usepackage{siunitx}
\usepackage{wrapfig}
\usepackage{amsthm}
\usepackage{url}            
\usepackage{booktabs}       
\usepackage{amsfonts}       
\usepackage{nicefrac}       
\usepackage{microtype}      
\usepackage{xcolor}         
\usepackage{thmtools}
\usepackage{thm-restate}
\usepackage[utf8]{inputenc} 
\usepackage[T1]{fontenc}    
\usepackage[capitalize]{cleveref}
\usepackage{hyperref}       
\usepackage{todonotes}
\usepackage{wasysym}
\usepackage[nonumberlist,acronym]{glossaries}

\makeglossaries
\newacronym{g0}{$G_{0}$}{Group 0}
\newacronym{g1}{$G_{1}$}{Group 1}
\newacronym{g0_inline}{G_{0}}{Group 0}
\newacronym{g1_inline}{G_{1}}{Group 1}
\newacronym{ml}{ML}{Machine Learning}

\pgfplotsset{width=6.8cm,compat=1.18}

\def\mytitle{Happiness as a Measure of Fairness}

\newtheorem{lemma}{Lemma}

\newtheorem{definition}{Definition}

\crefname{equation}{Equation}{Equations}
\Crefname{equation}{Equation}{Equations}
\crefformat{equation}{#2(#1)#3}
\crefrangeformat{equation}{#3(#1)#4--#5(#2)#6}
\crefmultiformat{equation}{#2(#1)#3}{ and~#2(#1)#3}{, #2(#1)#3}{ and~#2(#1)#3}
\Crefformat{equation}{#2(#1)#3}
\Crefrangeformat{equation}{#3(#1)#4--#5(#2)#6}
\Crefmultiformat{equation}{#2(#1)#3}{ and~#2(#1)#3}{, #2(#1)#3}{ and~#2(#1)#3}
\crefname{figure}{Figure}{Figures}
\Crefname{figure}{Figure}{Figures}

\def\happy{\boldsymbol{\eta}}
\def\ehappy{\boldsymbol{\xi}}
\def\fair{\boldsymbol{\phi}}
\def\eps{\varepsilon}
\def\ind{\mathbbm 1}

\begin{document}

%

%

\twocolumn[

  \conferencetitle{\mytitle}

  \conferenceauthor{ Georg Pichler\textsuperscript{\normalfont 1} \And Marco Romanelli\textsuperscript{\normalfont 2} \And  Pablo Piantanida\textsuperscript{\normalfont 3}}

 \conferenceaddress{
\textsuperscript{1}Institute of Telecommunications, TU Wien,
Vienna University of Technology \\
\textsuperscript{2}Computer Science, Hofstra University \\
\textsuperscript{3}Université Paris-Saclay, CNRS, Mila
} ]

\begin{abstract}
  In this paper, we propose a novel fairness framework grounded in the concept of \emph{happiness}, a measure of the utility each group gains from decision outcomes. By capturing fairness through this intuitive lens, we not only offer a more human-centered approach, but also one that is mathematically rigorous: In order to compute the optimal, fair post-processing strategy, only a linear program needs to be solved. This makes our method both efficient and scalable with existing optimization tools. Furthermore, it unifies and extends several well-known fairness definitions, and our empirical results highlight its practical strengths across diverse scenarios.

\end{abstract}

\section{\uppercase{Introduction}}
\label{sec:introduction}

In classification problems where the considered dataset can be naturally divided into several groups, such as by a specific attribute, it is often desirable for the classifier to treat each group ``equally''.
However, unfair results are often observed and may arise from a variety of sources, such as imbalances in the training set, biases introduced during the learning process, or existing biases in the training data itself. These unfair outcomes typically mean that groups are not treated equally, receiving systematically better or worse outcomes compared to others.

Many techniques have been developed to enhance fairness in \gls*{ml} systems, e.g., \citep{HardtPS2016NeurIPS, JiangHFYMH2022ICLR, AgarwalBDLW2018ICML, CatonH2024ACMCS, GoharC2023IJCAI}. These methods can be applied at different stages of the \gls*{ml} pipeline: pre-processing techniques adjust the training data before learning begins, in-processing methods guide the training to promote fair outcomes, and post-processing approaches modify the model's outputs to reduce unfairness after training is complete. In the latter case, the output of a soft-classifier is post-processed, ensuring fair classification in the process, typically at the expense of the overall accuracy.

Popular post-processing techniques aim to equalize statistical measures involving only labels and group membership \cite{BerkHJKR2021SMR,BhartiYS2023NeurIPS,HardtPS2016NeurIPS,JiangHFYMH2022ICLR,TangZ2022Causality}. In doing so, fairness is often framed through the lens of group-wise performance metrics (e.g., accuracy, false positive rates, or precision) valuated separately for each demographic. When these metrics are aligned across groups, the model is considered fair. However, this performance-centric perspective, while intuitive, can obscure deeper forms of unfairness particularly when the training data already embeds historical or structural biases. By focusing solely on output parity, these approaches risk masking disparities in how different groups experience the outcomes, leading to fairness definitions that are technically satisfied but practically insufficient. Our work challenges this narrow perspective and proposes an alternative one that captures the utility derived by each group, offering a richer and potentially more just notion of fairness.

To illustrate the limitations of current fairness metrics, consider a \gls*{ml} system designed to determine whether to grant a loan to an individual. The input $X$ includes features such as the borrower's profile and the amount of credit requested, and the output $Y$ is a binary decision indicating loan approval or rejection. Now suppose we have two demographic groups, \gls*{g0} and \gls*{g1}, that are identical in all respects except that, on average, individuals from \gls*{g0} request twice as much credit as those from \gls*{g1}. A classifier that approves loans at the same rate for both groups, thus appearing fair under standard group-based metrics, would in fact result in an unequal allocation of resources: group \gls*{g1} would receive significantly less total credit than group \gls*{g0}, despite the groups being otherwise indistinguishable. This outcome is intuitively unfair, but standard fairness definitions based on performance metrics (e.g., Equalized Odds or demographic parity) would fail to capture it, since they only consider prediction correctness or distribution, not the magnitude or utility of the outcome. Moreover, if this imbalance is already embedded in the training data, the classifier may simply learn to replicate it, and fairness constraints applied post hoc will not rectify the core issue. 


\subsection{Contributions}
\label{sec:contributions}
\begin{itemize}

  \item We introduce the concept of \textbf{happiness} as a tool
        reflecting how satisfied an individual is with the output of a classifier. Fairness can then be measure by how evenly happiness is allocated between groups--where perfect equality in allocation corresponds to equal happiness across groups (cf.\ \cref{sec:measuring-fairness}), e.g. the loan amount in the example above.
  \item We apply our definition to find the optimal post-processing strategy, resulting in a linear program which can be solved efficiently (cf.\ \cref{sec:fairness}).
  \item Importantly, we show how this definition encompasses other popular definitions of fairness as special cases (cf.\ \cref{sec:recov-other-crit}).
  \item We provide numerical experiments showcasing the utility of this approach (cf.\ \cref{sec:caseStudies}).

\end{itemize}
\subsection{Related Work}
\label{sec:related-work}
The problem of data bias and its impact on the fairness of \gls*{ml} models is well documented in the literature \cite{BarocasHN2023FairML,CerratoKWK2024CoRR}. Over the years, alongside the development of dedicated datasets for fairness evaluation \cite{LeQuyRIZN2022DMKD}, three main categories of approaches have been proposed to mitigate bias in \gls*{ml} models: (i) pre-processing approaches, which aim to modify the training data to reduce or eliminate bias; (ii) in-processing approaches, which seek to adjust the learning algorithm itself to address biases caused by dominant features or other distributional effects; and (iii) post-processing approaches, which apply transformations to the model’s output to enhance fairness in predictions \citep{GoharC2023IJCAI,MehrabiMSLG2022ACMCS,FabrisMSS2022EEAMO,CatonH2024ACMCS}.

Post-processing methods are extremely popular in the literature due to their \emph{flexibility}: they do not explicitly modify the underlying model and therefore do not require access to the training algorithms or the models themselves. They are also valued for their \emph{lightweight} nature, as they only require access to the model’s predictions and sensitive attribute information. Furthermore, their \emph{ease of use} makes them appealing in scenarios where modifying the data or the model may have legal implications or compromise their interpretability.

Several definitions of fairness, all linked to post-processing approaches that modify model decisions to optimize specific metrics, have been proposed in the literature. Among the most prominent are \emph{Overall Accuracy Equality} \cite{BerkHJKR2021SMR}, \emph{Equalized Odds} \cite{HardtPS2016NeurIPS,BhartiYS2023NeurIPS,TangZ2022Causality} and \emph{Equal Opportunity} \cite{HardtPS2016NeurIPS}, \emph{Demographic Parity} \cite{JiangHFYMH2022ICLR}, and \emph{Statistical Parity} \cite{DworkHPRZ2012ITCSC}, to name a few. These fairness criteria guide prediction correction by balancing
the distribution of predicted labels with respect to the sensitive attributes which characterize different groups \cite{VermaR2018IWSF}.

A variety of extensions have been proposed to generalize these definitions to multi-class problems \cite{AlghamdiHJM2022NeurIPS,LiuWWWSG2023AAAI,Rouzot2023Learning}, multi-group settings \cite{DworkLLT2023CoRR}, and even regression tasks \cite{TaturyanCH2024NeurIPS}. Other lines of work have explored entirely different directions, such as leveraging the theory of calibration \cite{PleissRWKW2017NeurIPS}, applying tools from optimal transport \cite{GordalizaBGL2019ICML,ChiappaRST2020AAAI,BuylD2022NeurIPS,WangHGC2023NeurIPS}, or using cryptographic primitives \cite{YadavRBC2024NeurIPS}. Perhaps more aligned with the work presented in this paper, \cite{AgarwalBDLW2018ICML,LiuWWWSG2023AAAI,WoodworthGOS2017COLT,LiuSH2019ICML,KimRR201018NeurIPS,PerdomoZMH2020ICML,ZafarVGG2019JMLR} offer various attempts at establishing a unified framework to obtain and or evaluate fairness.
However, a key distinction lies in the fact that none of these approaches are equipped to incorporate the notion of happiness introduced in this work. As a result, they are not well-suited for post-processing settings where capturing individual or group-level utility is essential.

More closely aligned with our framework are the works of \cite{liu2018delayed,weber2022enforcing}, which introduce the notion of delayed impact and long-term fairness effects. These approaches are orthogonal to ours, as they examine how conventional fairness constraints on decision policies affect long-term outcomes. To this end, repeated classification is considered where the classifier optimizes a utility function tied to institutional gain, such as load repayment probability for a bank or loan office. In these settings, fairness impacts emerge through changes in external variables rather than the label itself. A complementary perspective is offered in \cite{kasy2021fairness}, where fairness constraints are modeled as institutional costs, particularly when treatment of a subpopulation is boosted to ensure merit-based recognition. This can introduce new disparities, whose impact can be characterized by influence functions, which are not captured by standard fairness metrics.
These works require a definition, quantifying how well a group performs at each point in time. Commonly, average credit score, when dealing with loan approval or educational success metrics when using admissions data, are used. This is akin to the definition of ``happiness'' introduced in this work. While our work does not include temporal modeling, in essence we propose to directly use the relevant metric when defining the fairness of a classifier. This is also mentioned in \cite[Sec.~4.3]{liu2018delayed}, but not expanded upon.

Thus, our framework introduces a general class of fairness constraints that subsume common definitions, while even enabling the optimization of long-term effects, within a setting where classification accuracy remains the core objective, relevant to both model developers and end users.


\section{\uppercase{Happiness as a Criterion for Evaluating Fairness}}
\label{sec:measuring-fairness}
While the scenario discussed in \cref{sec:introduction}, involving two identical groups where unfair behavior arises from biased training data, is intentionally contrived, it showcases the need for a framework to address the issue of disparate resource allocation. In this paper, we focus on a post-processing strategy that can be applied to any soft classifier, resulting in a fair classification. Our method naturally allows for a trade-off between accuracy and fairness. In the following, we will provide a brief outline of our method and its advantages.

Owing to the fact that performance metrics alone can be insufficient for determining fairness, for the first time, we take a more general approach and define a \emph{happiness function} $\happy$, which, in general, takes all features, the group index characterized by the sensitive feature(s), the ground truth as well as the (hard) classifier output label as inputs. It produces a real number as output, which quantifies the happiness of an individual with the classifier output.
If -- on average -- the happiness of individuals in the two groups are close, a classifier is considered fair. Note that the happiness can really be an arbitrary function of all features, and
$\happy$ is only required to output a real number.

The choice of the happiness function must be tailored to the specific problem setting and the intended use of the classifier. Notably, the same classifier may require distinct happiness formulations depending on its application context. For example, consider a classifier trained to estimate (or quantize) individual income. The perceived utility (or \emph{happiness}) associated with a given prediction would naturally differ depending on whether the system is employed for assessing credit risk or determining taxation. This highlights the importance of aligning the happiness function with the operational goals and stakeholder perspectives inherent to each use case.


\subsection{Proposed Framework}
\label{sec:our-algorithm}
We restrict ourselves to linear, group-dependent post-processing of soft classifiers. Thus, our post-processing step can be described by a conditional probability distribution on the finite label space for each group.

The resulting trade-off between classification accuracy and fairness naturally yields a linear optimization problem: the accuracy of the classification is a linear function of these conditional distributions. The expected happiness (conditioned on the group) is also a linear function of the conditional post-processing probability distributions. This holds for arbitrary happiness functions. Thus, finding the optimal post-processing rule is equivalent to solving a linear program. Our method is introduced in detail in \cref{sec:fairness}.

It was already pointed out in \cite{AgarwalBDLW2018ICML} that many criteria for fairness can be phrased as linear constraints. We can use this fact and recover these fairness notions using a happiness function if we allow vector-valued $\happy$. This does not create any complications in the optimization procedure, as it merely introduces additional linear constraints. In \cref{sec:recov-other-crit} we formally show, how $\happy$ has to be chosen to recover ``Statistical parity'', ``Overall accuracy'' and ``Equalized odds'' from our definitions, showcasing the flexibility and generality of the proposed framework.

A major advantage of defining \emph{fairness in terms of equal happiness of all groups}, is that it can be \emph{tailored towards the particular application} of the classification system. It also allows for \emph{additional features beyond the classification result and ground-truth} to be used in the computation of happiness. In case of a system designed for credit approval, e.g., the size of the loan can also be considered in addition to the binary decision of whether to approve the loan. Yet, our method retains the advantageous property that the computational problem is still a linear program.

Although our methods can be readily extended to accommodate more than two groups introducing additional constraints, such an extension is not pursued in the present work in the interest of brevity and clarity.

\subsection{Illustrative Example}
\label{sec:example}
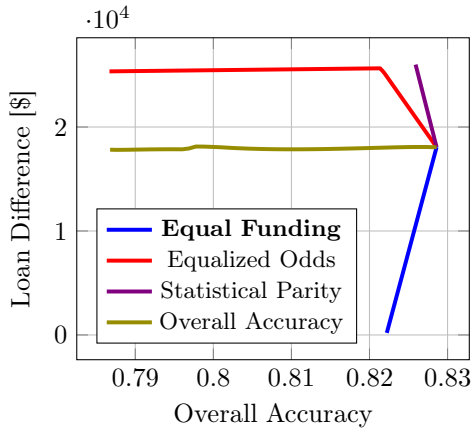
\begin{figure}[!htb]
  \centering
  \begin{tikzpicture}
  \begin{axis}[
    grid,
    legend entries={\textbf{Equal Funding}, Equalized Odds, Statistical Parity, Overall Accuracy},
    legend style={
      at={(.05,.05)},
      anchor=south west,
      fill=white,
      font=\small,
      legend image post style={line width=1.5pt}
    },
    ylabel={Loan Difference [\$]},
    xlabel={Overall Accuracy}
    ]
    \tikzset{every plot/.append style={mark=none, line width = 1.5pt}}
    \addplot+ table [x=acc, y=0, col sep=comma] {./plots/synthetic/acc_vs_happiness_delta_test.csv};
    \addplot+ table [x=acc, y=1, col sep=comma] {./plots/synthetic/acc_vs_happiness_delta_test.csv};
    \addplot[color=violet] table [x=acc, y=2, col sep=comma] {./plots/synthetic/acc_vs_happiness_delta_test.csv};
    \addplot[color=olive] table [x=acc, y=3, col sep=comma] {./plots/synthetic/acc_vs_happiness_delta_test.csv};
    \end{axis}
\end{tikzpicture}
  \caption{Our post-processing method (Equal Funding) guarantees any target accuracy level (up to 83.3\%) while minimizing funding disparities between groups. Notably, it achieves perfect fairness, i.e., zero difference in loan allocations between \gls*{g0} and \gls*{g1} with less than a one percentage point loss in accuracy w.r.t.\ the baseline unfair classifier.\vspace*{-15pt}}
  \label{fig:intro_plot}%
\end{figure}%
To motivate our method, we provide an example, comparing it to other methods of post-processing with different fairness metrics.
In this example, an \gls*{ml} system is tasked with approving (or denying) a loan. We use a completely synthetic dataset inspired by \textsc{Adult}~\citep{adult_2}, divided in two groups, namely \gls*{g0} and \gls*{g1}. All features are independently generated. The annual income is drawn from the Gaussian distribution $\mathcal(\mu, \sigma^2)$ with mean $\mu_0 = \SI{50000}[\$]{}$ and standard deviation $\sigma_0 = \SI{1000}[\$]{}$. The \emph{base} loan amount is an independently drawn Gaussian with standard deviation $\SI{10000}[\$]{}$, and mean $\mu_1 = \SI{500000}[\$]{}$. A loan is granted if the loan amount is less or equal than 10 times the annual income. Thus, a loan is granted with a probability of $0.5$. However, \gls*{g0} requests the base loan amount, while for \gls*{g1}, $\SI{50000}[\$]{}$ are added to the loan, while the probability of approval remains the same. This results in the allocation of additional funds to \gls*{g1} in the training data.

Subsequently, we train a random forest classifier on this linear classification problem, achieving an expected soft accuracy of $0.97$ on the training set and $0.833$ on the test set. Given any accuracy ${\alpha \le 0.833}$, a post-processing strategy is applied to the classifier, that, while guaranteeing accuracy $\alpha$ maximizes fairness. E.g., in the case of Overall Accuracy~\cite{BerkHJKR2021SMR} this procedure minimizes the absolute difference between the accuracy on the two groups, while maximizing overall accuracy.
For each value $\alpha$ we then compute the resulting difference in funding allocated to the two groups on average, when this post-processing step is applied. The plot in \cref{fig:intro_plot} shows the difference in loan amount as a function of accuracy $\alpha$.

Note that our method, dubbed ``Equal funding'' for this specific application, allows us to directly constrain the difference in funding while maintaining high accuracy.
It is noteworthy that the other three methods, in this case, never achieve a meaningful reduction of the difference between the two groups. At best, in the case of Overall Accuracy, the difference remains unchanged, while optimization for Equalized Odds and Statistical Parity even increase the imbalance between the groups.

In the context of \cref{fig:intro_plot}, it is worth to point out, that the Equalized Odds predictor as defined in \cite{HardtPS2016NeurIPS} corresponds to a single point in \cref{fig:intro_plot}: it is the left-most point of the ``Equalized Odds'' line, where equal odds for both groups correspond to about $0.795$ accuracy and a funding gap of \SI{25384}[\$]{}.

\section{\uppercase{Main Definitions and Theoretical Results}}
\label{sec:fairness}

We consider a standard classification problem, where $X \in \mathcal X = \mathbb R^d$ is the (random) feature vector, $Y \in \mathcal Y$ is the label in a finite space $\mathcal Y$, and
in addition, we use $Z \in \{0,1\}$ to denote the group characterized by the sensitive feature(s). 


A \textbf{soft classifier} is a function of $\hat Y(X)$, taking the features $X$ as input and producing a probability distribution on the labels $\mathcal Y$ as its output.
We can interpret such a classifier as a random variable $\hat Y$ which depends on $(Y,Z)$ only through $X$, i.e., $(Y,Z) - X - \hat Y$ form a Markov chain. This reflects the fact that in general, the classifier does not have access to $Y$ and $Z$ directly.
We interpret $\hat Y$ as a (random) estimator, estimating $Y$ from $X$.
Performance of the estimator is judged by a \textbf{loss function}. For simplicity, we will use the probability of incorrect classification:
\begin{align}
  \label{eq:ell}
  \ell(Y, \hat Y) := \mathbb P\{Y \neq \hat Y\} = \mathbb E[\ind\{Y \neq \hat Y\}] .
\end{align}
However, the results of this paper also hold for other loss functions.

Given an individual in group $z$ with features $x$ and true label $y$, we seek to quantify how happy they are with a classification results $\hat y$. This notion is captured by a \textbf{happiness} function $\happy\colon \mathcal Y \times \mathcal X \times \mathcal Y \times \mathcal Z \to \mathbb R^n$, where $n \in \mathbb N$, which gives the happiness score $\happy(\hat y, x, y, z)$.
When $n > 1$, happiness is simply measured by multiple scalar happiness functions simultaneously.

Given a trained estimator $\hat Y$, we want to perform demography-dependent post-processing, increasing the fairness of the estimator, while maintaining high accuracy.
This can be achieve by another soft classifier $\tilde{Y}$, which is given $(\hat Y, Z)$, the (random) output of the original estimator and the group index. The soft classifier $\tilde Y$ produces another probability distribution at its output.
Thus, $\tilde Y$ can be interpreted as another random variable, which depends on all other random variables only through $\hat Y$ and $Z$, i.e., it is demography-dependent and we have the Markov chain $(X,Y) - (\hat Y, Z) - \tilde{Y}$.
This leads to the following factorization of the complete probability distribution:
\begin{align}
  p_{XYZ \hat{Y} \tilde{Y}}(x,y,z,\hat{y},\tilde{y}) & = p_{XYZ}(x,y,z) p_{\hat{Y}|XYZ}(\hat{y}|x,y,z) \nonumber                                                                       \\*
                                                     & \qquad\cdot p_{\tilde{Y}|\hat{Y}XYZ}(\tilde{y}|\hat{y},x,y,z)                                                                   \\
                                                     & \hspace{-60pt} = p_{XYZ}(x,y,z) p_{\hat{Y}|X}(\hat{y}|x) p_{\tilde{Y}|\hat{Y},Z}(\tilde{y}|\hat{y},z) . \label{eq:markov_chain}
\end{align}

To complete the problem setup, we need to define fairness in terms of happiness and specify the trade-off between fairness and accuracy resulting from that definition.
We say that a classifier is $\eps$-fair if the average happiness of individuals in \gls*{g0} and \gls*{g1} are no more than $\eps$ apart. This results in a trade-off between fairness and accuracy, where we are interested in minimizing the loss $\ell(Y, \tilde Y)$ among all $\eps$-fair classifiers $\tilde Y$ for some fixed $\eps > 0$.
This is formalized in the following definition.
\begin{definition}
  \label{def:main}
  For a happiness function $\happy$ and $ \eps\ge 0$, an estimator $\hat Y$ is \textbf{$\eps$-fair} if\footnote{For a vector $\mathbf{a} = (a_1, a_2, \dots a_n)$ and a scalar $b$, we use the notation $\mathbf{a} \le b$ for component-wise inequality $a_i \le b$ for all $i \in \{1,\dots,n\}$.}
  \begin{align}
    \label{eq:fair}
    \fair(\happy, \hat Y) & := \big| \mathbb E[\happy(\hat Y, X, Y, Z)|Z = 0] \nonumber             \\*
                          & \qquad\qquad- \mathbb E[\happy(\hat Y, X, Y, Z)|Z = 1] \big| \le \eps .
  \end{align}

  For a happiness function $\happy$ and a soft classifier $\hat Y$, a pair $(\eps, L)$ is \textbf{achievable} if there exists a demographic-dependent post-processing, i.e., an estimator $\tilde{Y}$ satisfying \cref{eq:markov_chain}, which is $\eps$-fair and
  satisfies $\ell(Y, \tilde{Y}) \le L$.
\end{definition}

Fortunately, though $\happy$ may depend on arbitrary features, the resulting problem is still a linear program.
\begin{restatable}{theorem}{main}
  \label{thm:main}
  For fixed $\eps \ge 0$ we can find the minimum $L$, such that $(\eps, L)$ is achievable by solving
  \begin{align}
    \min_{p_{\tilde{Y}|\hat YZ}} & \ell(Y, \tilde{Y})                                                                    \label{eq:main:1} \\
    \text{s.t. }                 & \fair(\happy, \tilde{Y}) \le \eps. \label{eq:main:2}
  \end{align}
  This is a linear programming problem.
\end{restatable}
The proof and further discussion of this result can be found in \cref{sec:proof-disc-main}.

In practice, however, the joint distribution of $(X, Y, Z)$ will not be available. Thus, we need to replace the expectations by empirical averages.
However, this will be reasonably accurate, even with modest training set sizes, as only the expected value of $2(n+1)|\mathcal Y|^2$ random variables needs to be approximated. Note that $n$ is the dimension of the happiness function $\happy$. Details are given in \cref{sec:empir-appr}, in particular \cref{lem:approx2}.

\section{\uppercase{Recovering Other Fairness Criteria}}
\label{sec:recov-other-crit}

In this section, we show that many previously established criteria of fairness can be recovered using the setup presented in \cref{sec:fairness}. We will showcase this using ``Statistical Parity'', ``Overall Accuracy'' and ``Equalized Odds''. We take advantage of the fact that these definitions can be phrased as linear constraints, which was already observed in a different context in \cite{AgarwalBDLW2018ICML}.

The definitions are taken from \cite[Table~II]{Rouzot2023Learning}. While the original definitions ask for exact equality, we will include $\eps \ge 0$. Thus, the original definitions from \cite[Table~II]{Rouzot2023Learning} can be recovered by setting $\eps = 0$ in what follows.

\begin{definition}[Statistical Parity]
  \label{def:stat_parity}
  A classifier $\hat Y$ has $\eps$ \emph{Statistical Parity} (or \emph{demographic parity}) if
  \begin{align}
    \label{eq:stat-parity}
    \big|p_{\hat Y|Z}(\cdot|0) - p_{\hat Y|Z}(\cdot|1)\big| \le \eps .
  \end{align}
\end{definition}
\begin{lemma}
  \label{lem:stat_parity}
  A classifier $\hat Y$ has $\eps$ Statistical Parity if and only if it is $\eps$-fair w.r.t.\ the $n = |\mathcal Y|$ dimensional happiness function $\happy(y,\hat y,z) = (\ind_{\tilde{y}}(\hat y))_{\tilde{y} \in \mathcal Y}$.
\end{lemma}
\begin{proof}
  Substituting the given happiness function $\happy$ in \cref{eq:fair},
  $\hat Y$ is $\eps$-fair if for all $\tilde{y} \in \mathcal Y$ we have
  \begin{align}
    \big| \mathbb E[\ind_{\tilde{y}}(\hat Y)|Z=0] - \mathbb E[\ind_{\tilde{y}}(\hat Y)|Z=1]\big| \le \eps ,
  \end{align}
  which is equivalent to \cref{eq:stat-parity}.
\end{proof}

\begin{definition}[Overall Accuracy]
  \label{def:overall-accuracy}
  A classifier $\hat Y$ has $\eps$ \emph{equal Overall Accuracy} if
  \begin{align}
    \label{eq:overall-accuracy}
    \big|P(Y = \hat Y|Z=0) - P(Y = \hat Y|Z=1)\big| \le \eps .
  \end{align}
\end{definition}
\begin{lemma}
  \label{lem:overall-accuracy}
  A classifier $\hat Y$ has $\eps$ equal Overall Accuracy if and only if it is $\eps$-fair w.r.t.\ the $n = 1$ dimensional happiness function $\happy(y,\hat y,z) = \ind_{\hat y}(y)$.
\end{lemma}
\begin{proof}
  The result follows from substituting the given happiness function in \cref{eq:fair}.
\end{proof}

\begin{definition}[Equalized Odds]
  \label{def:eq_odds}
  A classifier $\hat Y$ has $\eps$ \emph{Equalized Odds} if for all $y \in \mathcal Y$,
  \begin{align}
    \label{eq:eq_odds}
    \big|p_{\hat Y|ZY}(\cdot|0, y) - p_{\hat Y|ZY}(\cdot|1,y)\big| \le \eps .
  \end{align}
\end{definition}
\begin{lemma}
  \label{lem:eq_odds}
  A classifier $\hat Y$ has $\eps$ Equalized Odds if and only if it is $\eps$-fair w.r.t.\ the $n = |\mathcal Y|^2$ dimensional happiness function $\happy(y,\hat y,z) = \left(\frac{\ind_{y', \tilde{y}}(y, \hat y)}{p_{Y|Z}(y'|z)}\right)_{y', \tilde{y} \in \mathcal Y}$.
\end{lemma}
\begin{proof}
  Note that
  \begin{align}
    \label{eq:eq_odds2}
    &\mathbb E[\happy_{y', \tilde{y} }(Y,\hat Y,z)|Z=z] = \frac{\mathbb E[\ind_{y', \tilde{y}}(Y, \hat Y)|Z=z]}{p_{Y|Z}(y'|z)} \\
                                                       &\qquad = \frac{p_{\hat Y Y|Z}(\tilde{y}, y'|z)}{p_{Y|Z}(y'|z)} 
                                                       = p_{\hat Y | Y Z}(\tilde{y}| y', z) .
  \end{align}
  Thus, substituting the given happiness function in \cref{eq:fair} is equivalent to \cref{eq:eq_odds} for all $y \in \mathcal Y$.
\end{proof}


\section{\uppercase{Demonstrative Case Studies}}
\label{sec:caseStudies}
In this section we report three studies which showcase the application of the proposed post-processing framework using the synthetic dataset mentioned in \cref{sec:example}, the \texttt{Adult} dataset \cite{adult_2}, and the \texttt{Financial Risk for Loan Approval} dataset \citep{financialRiskData}, respectively. All our experiments in this section follow the same procedure, which we will describe in the following.

\paragraph*{Dataset and Baseline Classifier.} All datasets contain a total of \SI{48842}{} samples. Given a dataset with features $X$, labels $Y$, and group labels $Z$, we split the dataset using a fixed random seed into training, validation, and test sets, containing $20\%$, $16\%$, and $64\%$ of the data, respectively.
We then train a simple random forest baseline classifier $\hat Y$ on the training data with accuracy $1 - \ell(Y, \hat Y)$.

\paragraph*{Outline of Experiments.} We define a custom happiness function $\happy$ and solve the linear program \cref{eq:main:1} for different values $\eps > 0$. For each $\eps$, we thereby obtain a $\eps$-fair classifier $\tilde Y$ with accuracy $A(\eps) = 1 - \ell(Y, \tilde Y)$. Note that $A(\eps)$ is monotonically increasing, thus $\eps(A)$ exists.
For comparison, we also post-process $\hat Y$ using ``Statistical Parity'', ``Overall Accuracy'' and ``Equalized Odds''. This can be achieved within our framework, but repeating the same process, replacing $\happy$ with a happiness function $\happy'$ for the corresponding method introduced in~\cref{sec:recov-other-crit}. Each method yields a family of post-processed classifiers $\tilde Y'$, for $\eps' > 0$. We report the difference in happiness $\eps(A') = \fair(\happy, \tilde Y')$ between the two groups measured by $\happy$, as a function of accuracy $A' = 1 - \ell(Y, \tilde Y')$, where $\ell(\cdot,\cdot)$ is defined in~\cref{eq:ell}.

For all classifiers, we perform the minimization~\cref{eq:main:1} by using the validation set for the empirical approximation of the expectation operator, as outlined in~\cref{sec:empir-appr}. The values $\eps(A)$, i.e., the happiness gap as a function of accuracy, is subsequently computed using the test and validation sets. Statistical robustness is evidenced by the fact that the results for the test and validation set are close.

\paragraph*{Computational Resources.} Each experiment was performed in less than three minutes on a \emph{AMD Ryzen 7 5700X} without GPU support. The memory required was less than $\SI{1}{GB}$. This includes dataset generation and training of the random forest classifier.
\subsection{Case Study with Synthetic Data}
\label{sec:toy_example}
In this \lcnamecref{sec:toy_example}, we provide a detailed explanation of the experiment introduced in~\cref{sec:example}.

We apply our method to a synthetic dataset, inspired by the \texttt{Adult} dataset~\cite{adult_2}.
All data in this dataset is randomly drawn. For the column $Z = X_{\texttt{sex}}$, which we use as group label, we keep the original imbalance of about 1/3 female and 2/3 male from \texttt{Adult}.
Values for the following features are drawn uniformly, independently at random: $X_{\texttt{age}}$, $X_{\texttt{hours\_per\_week}}$, $X_{\texttt{education}}$, $X_{\texttt{workclass}}$ and $X_{\texttt{race}}$. These features are completely independent of the classification task. They merely ensure that the linear classification problem introduced next is not learned perfectly by the baseline classifier.

Values for a new feature $X_{ \texttt{yearly\_salary}}$ are normally distributed with mean $\mu_0 = \SI{50000}[\$]{}$ and standard deviation $\sigma_0 = \SI{1000}[\$]{}$. The \emph{base} loan amount $U$ is an independently drawn Gaussian with standard deviation $\SI{10000}[\$]{}$, and mean $\mu_1 = \SI{500000}[\$]{}$. A loan is granted if the \emph{base} loan amount is less or equal than 10 times the annual income, i.e., $Y = \ind\{10\cdot X_{ \texttt{yearly\_salary}} \ge U\}$. Thus, a loan is granted with a probability of $0.5$. However, while male applicants request the base loan amount $X_{\texttt{loan\_requested}} = U$, for female applicants, $\SI{50000}[\$]{}$ are added to the loan, $X_{\texttt{loan\_requested}} = U + \SI{50000}[\$]{}$. The decision of acceptance is based on the \emph{base} loan amount $U$, skewing the approval and resulting in the allocation additional funds to female applicants. This bias is already present in the training data.

Our baseline classifier is a simple random forest model trained on the training set, achieving approximately $82\%$ accuracy on the test set.

The considered happiness function, based on the prediction and the requested loan, is defined as
\begin{equation*} \happy(\hat Y, X, Y, Z) = \hat Y \cdot X_{\texttt{loan\_requested}}, \end{equation*}
where happiness is set to zero for rejected loan requests and to the loan amount for approved ones. Although simplistic, this example effectively illustrates the flexibility of our proposed framework. It demonstrates that ensuring equalized false and true positive rates across groups does not necessarily guarantee fairness in the allocation of resources.


In \cref{fig:intro_plot}, we observe optimizing for any of ``Statistical Parity'', ``Overall Accuracy'' and ``Equalized Odds'' does not improve fairness in terms of allocated funding measured by $\happy$.
Indeed ``Statistical Parity'' and ``Equalized Odds'' even amplify the bias in the training data, leading to an even larger difference after post-processing.
All methods maintain reasonable classification performance, but only directly minimizing the difference in funding allows for a reduction of the imbalances in monetary allocation across groups. Note that this is possible while sacrificing less than one percentage point in accuracy.
\Cref{fig:intro_plot} shows the test set results; the validation set plots are in \cref{sec:details:synthetic}.

\subsection{Case Study with \texttt{Adult} Data}
In this case study, we focus on the \texttt{Adult} dataset \citep{adult_2}. While we retain the standard classification task of predicting whether an individual's income exceeds \SI{50000}[\$]{}, i.e., the hard decision is $1$ if the predicted income is at least \SI{50000}[\$]{} and $0$ otherwise. We expand the analysis to incorporate a broader perspective on individual well-being.

Specifically, we use our framework to highlight that an individual who earns at least \SI{50000}[\$]{} without working excessive hours may experience greater happiness. We posit that holding a job that yields higher income for fewer hours per week contributes positively to overall well-being, as the saved time can be reallocated to additional income-generating activities or personal pursuits.

To formalize this intuition, we define a happiness function as 
$
  \happy(\hat{Y}, X, Y, Z) = 100 \cdot \hat{Y} - X_{\texttt{hours\_per\_week}}
$,
where $\hat{Y}$ is the predicted label, and the feature $X_{\texttt{hours\_per\_week}}$, indicating the number of hours an individual works each week has support is $\mathcal{X}_{\texttt{hours\_per\_week}} = \{ x \in \mathbb{Z} \mid 1 \leq x \leq 99 \}
$. This function captures the idea that achieving a high income with fewer working hours leads to increased happiness.

\label{sec:example_adultdata}
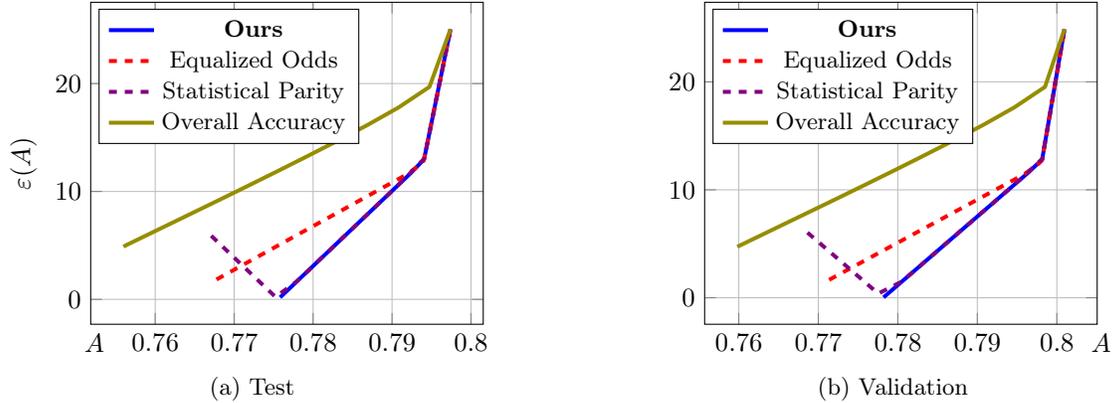
\begin{figure*}[!htb]
  \centering
  \begin{subfigure}{0.49\linewidth}
    \centering
    \begin{tikzpicture}
  \begin{axis}[
    grid,
    legend entries={\textbf{Ours}, Equalized Odds, Statistical Parity, Overall Accuracy},
    legend style={
      at={(.02,.98)},
      anchor=north west,
      fill=white,
      font=\small,
      legend image post style={line width=1.5pt}
    },
    x label style={at={(axis description cs:0.01,0.0)},anchor=north},
    ylabel={$\eps(A)$},
    xlabel={$A$},
    ]
    \tikzset{every plot/.append style={mark=none, line width = 1.5pt}}
    \addplot+[opacity=0.6] table [x=acc, y=0, col sep=comma] {./plots/adult/acc_vs_happiness_delta_test.csv};
    \addplot+[opacity=0.5, dashed] table [x=acc, y=1, col sep=comma] {./plots/adult/acc_vs_happiness_delta_test.csv};
    \addplot[color=violet, opacity=0.5, dashed] table [x=acc, y=2, col sep=comma] {./plots/adult/acc_vs_happiness_delta_test.csv};
    \addplot[color=olive, opacity=0.5] table [x=acc, y=3, col sep=comma] {./plots/adult/acc_vs_happiness_delta_test.csv};
    \end{axis}
\end{tikzpicture}
    \caption{Test}
    \label{fig:adult:test}
  \end{subfigure}
  \begin{subfigure}{0.49\linewidth}
    \centering
    \begin{tikzpicture}
  \begin{axis}[
    grid,
    legend entries={\textbf{Ours}, Equalized Odds, Statistical Parity, Overall Accuracy},
    legend style={
      at={(.02,.98)},
      anchor=north west,
      fill=white,
      font=\small,
      legend image post style={line width=1.5pt}
    },
    x label style={at={(axis description cs:1.01,0.0)},anchor=north},
    xlabel={$A$},
    ]
    \tikzset{every plot/.append style={mark=none, line width = 1.5pt}}
    \addplot+[opacity=0.6] table [x=acc, y=0, col sep=comma] {./plots/adult/acc_vs_happiness_delta_val.csv};
    \addplot+[opacity=0.5, dashed] table [x=acc, y=1, col sep=comma] {./plots/adult/acc_vs_happiness_delta_val.csv};
    \addplot[color=violet, opacity=0.5, dashed] table [x=acc, y=2, col sep=comma] {./plots/adult/acc_vs_happiness_delta_val.csv};
    \addplot[color=olive, opacity=0.5] table [x=acc, y=3, col sep=comma] {./plots/adult/acc_vs_happiness_delta_val.csv};
    \end{axis}
\end{tikzpicture}
    \caption{Validation}
    \label{fig:adult:val}
  \end{subfigure}
  \caption{Experiment on \texttt{Adult} dataset.}
  \label{fig:adult}
\end{figure*}

Consistent with the observations reported in \cref{sec:toy_example}, \cref{fig:adult} illustrates that our post-processing method achieves the most favorable trade-off between accuracy and happiness when the decision-making process accounts for the average number of working hours across different demographic groups. While the scenario depicted is simplified, it is representative of practical settings where one may wish to deploy our method to promote equitable working conditions for different groups. For instance, it could be applied to foster a work environment in which no group is disproportionately required to work overtime in order to receive comparable benefits, here represented by earning a salary above or below \SI{50000}[\$]{}. More broadly, our approach may be viewed as a scalable tool for addressing structural disparities in the workplace, with potential to mitigate issues such as high employee turnover.

Interestingly, while the ``Statistical Parity'' baseline achieves similar performance to our method in this particular case, it does not explicitly optimize for happiness. This limitation is reflected in the behavior of the growth function: the disparity in happiness between the two groups increases at both ends of the accuracy spectrum, with a localized point where the difference is negligible. \cref{fig:adult:val} further showcases how our proposed solution achieves minimal $\eps(A)$ on the validation dataset, in stark contrast with ``Statistical Parity'' and the other criteria.

\subsection{Case Study with \texttt{Financial Risk for Loan Approval} Data}
\label{sec:example_loan_risk}
The \texttt{Financial Risk for Loan Approval} dataset \citep{financialRiskData} is used in classification tasks aimed at identifying individuals who pose a low risk of defaulting on a mortgage. Unlike \texttt{Adult}, this dataset includes more detailed information about the applicant's financial and employment status, factors commonly used by financial institutions when evaluating loan applications. These include, among others, employment or unemployment status, duration in that status, credit score, and the stated reason for requesting a loan.

Beyond risk assessment, such indicators can also be leveraged to model a user's perceived utility or happiness as a tradeoff between securing a mortgage and managing the repayment burden under the constraints of their economic situation as understood by financial institutions. In particular, we define a happiness function that models the potential impact of loan approval decisions on an individual's well-being, under the assumption that a loan can lead to long-term financial gains (e.g., home ownership, business investment), but also comes with repayment obligations (i.e., future costs due to interest).

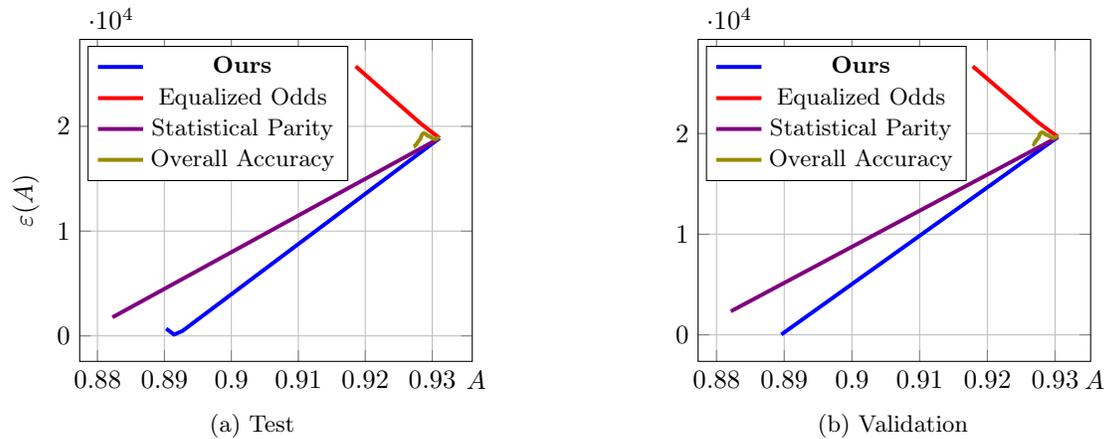
\begin{figure*}[!htb]
  \centering
  \begin{subfigure}{0.49\linewidth}
    \centering
    \begin{tikzpicture}
  \begin{axis}[
    grid,
    legend entries={\textbf{Ours}, Equalized Odds, Statistical Parity, Overall Accuracy},
    legend style={
      at={(.02,.98)},
      anchor=north west,
      fill=white,
      font=\small,
      legend image post style={line width=1.5pt}
    },
    x label style={at={(axis description cs:1.01,0.0)},anchor=north},
    ylabel={$\eps(A)$},
    xlabel={$A$},
    ]
    \tikzset{every plot/.append style={mark=none, line width = 1.5pt}}
    \addplot+[] table [x=acc, y=0, col sep=comma] {./plots/financial/acc_vs_happiness_delta_test.csv};
    \addplot+[] table [x=acc, y=1, col sep=comma] {./plots/financial/acc_vs_happiness_delta_test.csv};
    \addplot[color=violet] table [x=acc, y=2, col sep=comma] {./plots/financial/acc_vs_happiness_delta_test.csv};
    \addplot[color=olive] table [x=acc, y=3, col sep=comma] {./plots/financial/acc_vs_happiness_delta_test.csv};
    \end{axis}
\end{tikzpicture}
    \caption{Test}
    \label{fig:financial:test}
  \end{subfigure}
  \begin{subfigure}{0.49\linewidth}
    \centering
    \begin{tikzpicture}
  \begin{axis}[
    grid,
    legend entries={\textbf{Ours}, Equalized Odds, Statistical Parity, Overall Accuracy},
    legend style={
      at={(.02,.98)},
      anchor=north west,
      fill=white,
      font=\small,
      legend image post style={line width=1.5pt}
    },
    x label style={at={(axis description cs:1.01,0.0)},anchor=north},
    xlabel={$A$},
    ]
    \tikzset{every plot/.append style={mark=none, line width = 1.5pt}}
    \addplot+[] table [x=acc, y=0, col sep=comma] {./plots/financial/acc_vs_happiness_delta_val.csv};
    \addplot+[] table [x=acc, y=1, col sep=comma] {./plots/financial/acc_vs_happiness_delta_val.csv};
    \addplot[color=violet] table [x=acc, y=2, col sep=comma] {./plots/financial/acc_vs_happiness_delta_val.csv};
    \addplot[color=olive] table [x=acc, y=3, col sep=comma] {./plots/financial/acc_vs_happiness_delta_val.csv};
    \end{axis}
\end{tikzpicture}
    \caption{Validation}
    \label{fig:financial:val}
  \end{subfigure}
  \caption{Experiment on \texttt{Financial Risk for Loan Approval} dataset.}
  \label{fig:financial}
\end{figure*}

We define the happiness function as
$
  \happy(\hat{Y}, X, Y, Z) = \hat{Y} \cdot \left(X_{\texttt{loan\_requested}} \cdot R(X) - C(X) \right)
$,
where $C(X) = X_{\texttt{loan\_requested}} \cdot \rho(X_{\texttt{credit\_score}}) \cdot X_{\texttt{duration}}$ is the total interest cost; $\rho(X_{\texttt{credit\_score}})$ is the interest rate determined by the credit score using the step function reported in \cref{sec:financialRiskDetails}; $R(X)$ is the estimated return on investment as defined in \cref{sec:financialRiskDetails}.

The function outputs zero if the loan is not approved (\(\hat{Y} = 0\)). This utility-based formulation captures both the benefit of approval and the financial burden of repayment, enabling evaluations beyond predictive accuracy. 

\cref{fig:financial} reports the trade-off between the accuracy of the classifier and the happiness function defined above. In particular, similarly to the case shown in \cref{sec:example_adultdata}, our proposed solution achieves higher accuracy among all the criteria for minimal $\eps(A)$ (cf. \cref{fig:financial:val}). As previously disclosed, this is slightly different for the test data (cf. \cref{fig:financial:test}), where the optimal value for $\eps(A)$ is not achieved by any criterion, though ours achieves the best accuracy overall when $\eps(A)$ approaches $0$.

\section{\uppercase{Summary and Concluding Remarks}}
\label{sec:summary}
We have introduced a novel post-processing criterion for fairness in \gls*{ml}, through the lens of happiness, a measure of group satisfaction with a classifier's output. The proposed post-processing strategy is formulated as a linear program and thus, it allows for efficient computation while begin general enough to encompass many existing fairness criteria as particular instances.

Finally, we demonstrated the practicality of our approach through a series of case studies. Importantly, our method is readily extensible to fairness across multiple groups by incorporating additional constraints into the problem formulation.

This represents a novel perspective on the challenge of ensuring equal treatment of diverse populations, accounting not only for the classifier’s output, but also for how this output translates into individual happiness. For example, this applies to scenarios involving the allocation of resources across groups, highlighting just one facet of the broader impact our framework can have in promoting equitable outcomes.

\subsection*{Limitations}
Our framework, in line with comparable criteria, focuses on group-level fairness and does not account for individual happiness, which means that the interests of outliers or individuals whose preferences significantly diverge from the group may be overlooked.

Additionally, the derivation of our algorithm relies crucially on two assumptions: that the label space is finite and that the classifier outputs a soft prediction, i.e., a probability distribution over the label space. As a result, the current formulation is not directly applicable to regression problems or settings where predictions are continuous rather than categorical.

\clearpage
\bibliography{bibliography}

\begin{thebibliography}{}

\bibitem[Agarwal et~al., 2018]{AgarwalBDLW2018ICML}
Agarwal, A., Beygelzimer, A., Dud{\'{\i}}k, M., Langford, J., and Wallach,
  H.~M. (2018).
\newblock A reductions approach to fair classification.
\newblock In Dy, J.~G. and Krause, A., editors, {\em Proceedings of the 35th
  International Conference on Machine Learning, {ICML} 2018,
  Stockholmsm{\"{a}}ssan, Stockholm, Sweden, July 10-15, 2018}, volume~80 of
  {\em Proceedings of Machine Learning Research}, pages 60--69. {PMLR}.

\bibitem[Alghamdi et~al., 2022]{AlghamdiHJM2022NeurIPS}
Alghamdi, W., Hsu, H., Jeong, H., Wang, H., Michalak, P.~W., Asoodeh, S., and
  Calmon, F. (2022).
\newblock Beyond adult and {COMPAS}: Fair multi-class prediction via
  information projection.
\newblock In Oh, A.~H., Agarwal, A., Belgrave, D., and Cho, K., editors, {\em
  Advances in Neural Information Processing Systems}.

\bibitem[Barocas et~al., 2023]{BarocasHN2023FairML}
Barocas, S., Hardt, M., and Narayanan, A. (2023).
\newblock {\em Fairness and Machine Learning: Limitations and Opportunities}.
\newblock MIT Press.

\bibitem[Becker and Kohavi, 1996]{adult_2}
Becker, B. and Kohavi, R. (1996).
\newblock {Adult}.
\newblock UCI Machine Learning Repository.
\newblock {DOI}: https://doi.org/10.24432/C5XW20.

\bibitem[Berk et~al., 2021]{BerkHJKR2021SMR}
Berk, R., Heidari, H., Jabbari, S., Kearns, M., and Roth, A. (2021).
\newblock Fairness in criminal justice risk assessments: The state of the art.
\newblock {\em Sociological Methods \& Research}, 50(1):3--44.

\bibitem[Bharti et~al., 2023]{BhartiYS2023NeurIPS}
Bharti, B., Yi, P., and Sulam, J. (2023).
\newblock Estimating and controlling for equalized odds via sensitive attribute
  predictors.
\newblock {\em Advances in neural information processing systems},
  36:37173--37192.

\bibitem[Buyl and De~Bie, 2022]{BuylD2022NeurIPS}
Buyl, M. and De~Bie, T. (2022).
\newblock Optimal transport of classifiers to fairness.
\newblock {\em Advances in Neural Information Processing Systems},
  35:33728--33740.

\bibitem[Caton and Haas, 2024]{CatonH2024ACMCS}
Caton, S. and Haas, C. (2024).
\newblock Fairness in machine learning: A survey.
\newblock {\em ACM Comput. Surv.}, 56(7).

\bibitem[Cerrato et~al., 2024]{CerratoKWK2024CoRR}
Cerrato, M., Köppel, M., Wolf, P., and Kramer, S. (2024).
\newblock 10 years of fair representations: Challenges and opportunities.

\bibitem[Dwork et~al., 2012]{DworkHPRZ2012ITCSC}
Dwork, C., Hardt, M., Pitassi, T., Reingold, O., and Zemel, R. (2012).
\newblock Fairness through awareness.
\newblock In {\em Proceedings of the 3rd innovations in theoretical computer
  science conference}, pages 214--226.

\bibitem[Dwork et~al., 2023]{DworkLLT2023CoRR}
Dwork, C., Lee, D., Lin, H., and Tankala, P. (2023).
\newblock From pseudorandomness to multi-group fairness and back.

\bibitem[Fabris et~al., 2022]{FabrisMSS2022EEAMO}
Fabris, A., Messina, S., Silvello, G., and Susto, G.~A. (2022).
\newblock Tackling documentation debt: A survey on algorithmic fairness
  datasets.
\newblock EAAMO '22, New York, NY, USA. Association for Computing Machinery.

\bibitem[Gohar and Cheng, 2023]{GoharC2023IJCAI}
Gohar, U. and Cheng, L. (2023).
\newblock A survey on intersectional fairness in machine learning: Notions,
  mitigation, and challenges.
\newblock In {\em Proceedings of the Thirty-Second International Joint
  Conference on Artificial Intelligence, {IJCAI} 2023, 19th-25th August 2023,
  Macao, SAR, China}, pages 6619--6627. ijcai.org.

\bibitem[Gordaliza et~al., 2019]{GordalizaBGL2019ICML}
Gordaliza, P., Barrio, E.~D., Fabrice, G., and Loubes, J.-M. (2019).
\newblock Obtaining fairness using optimal transport theory.
\newblock In Chaudhuri, K. and Salakhutdinov, R., editors, {\em Proceedings of
  the 36th International Conference on Machine Learning}, volume~97 of {\em
  Proceedings of Machine Learning Research}, pages 2357--2365. PMLR.

\bibitem[Hardt et~al., 2016]{HardtPS2016NeurIPS}
Hardt, M., Price, E., and Srebro, N. (2016).
\newblock Equality of opportunity in supervised learning.
\newblock In Lee, D.~D., Sugiyama, M., von Luxburg, U., Guyon, I., and Garnett,
  R., editors, {\em Advances in Neural Information Processing Systems 29:
  Annual Conference on Neural Information Processing Systems 2016, December
  5-10, 2016, Barcelona, Spain}, pages 3315--3323.

\bibitem[Jiang et~al., 2022]{JiangHFYMH2022ICLR}
Jiang, Z., Han, X., Fan, C., Yang, F., Mostafavi, A., and Hu, X. (2022).
\newblock Generalized demographic parity for group fairness.
\newblock In {\em International Conference on Learning Representations}.

\bibitem[Kasy and Abebe, 2021]{kasy2021fairness}
Kasy, M. and Abebe, R. (2021).
\newblock Fairness, equality, and power in algorithmic decision-making.
\newblock In {\em Proceedings of the 2021 ACM conference on fairness,
  accountability, and transparency}, pages 576--586.

\bibitem[Kim et~al., 2018]{KimRR201018NeurIPS}
Kim, M., Reingold, O., and Rothblum, G. (2018).
\newblock Fairness through computationally-bounded awareness.
\newblock {\em Advances in neural information processing systems}, 31.

\bibitem[Le~Quy et~al., 2022]{LeQuyRIZN2022DMKD}
Le~Quy, T., Roy, A., Iosifidis, V., Zhang, W., and Ntoutsi, E. (2022).
\newblock A survey on datasets for fairness-aware machine learning.
\newblock {\em Wiley Interdisciplinary Reviews: Data Mining and Knowledge
  Discovery}, 12(3):e1452.

\bibitem[Liu et~al., 2018]{liu2018delayed}
Liu, L.~T., Dean, S., Rolf, E., Simchowitz, M., and Hardt, M. (2018).
\newblock Delayed impact of fair machine learning.
\newblock In {\em International Conference on Machine Learning}, pages
  3150--3158. PMLR.

\bibitem[Liu et~al., 2019]{LiuSH2019ICML}
Liu, L.~T., Simchowitz, M., and Hardt, M. (2019).
\newblock The implicit fairness criterion of unconstrained learning.
\newblock In {\em International Conference on Machine Learning}, pages
  4051--4060. PMLR.

\bibitem[Liu et~al., 2023]{LiuWWWSG2023AAAI}
Liu, T., Wang, H., Wang, Y., Wang, X., Su, L., and Gao, J. (2023).
\newblock Simfair: {A} unified framework for fairness-aware multi-label
  classification.
\newblock In Williams, B., Chen, Y., and Neville, J., editors, {\em
  Thirty-Seventh {AAAI} Conference on Artificial Intelligence}, pages
  14338--14346. {AAAI} Press.

\bibitem[Mehrabi et~al., 2021]{MehrabiMSLG2022ACMCS}
Mehrabi, N., Morstatter, F., Saxena, N., Lerman, K., and Galstyan, A. (2021).
\newblock A survey on bias and fairness in machine learning.
\newblock {\em ACM Comput. Surv.}, 54(6).

\bibitem[Perdomo et~al., 2020]{PerdomoZMH2020ICML}
Perdomo, J., Zrnic, T., Mendler-D{\"u}nner, C., and Hardt, M. (2020).
\newblock Performative prediction.
\newblock In {\em International Conference on Machine Learning}, pages
  7599--7609. PMLR.

\bibitem[Pleiss et~al., 2017]{PleissRWKW2017NeurIPS}
Pleiss, G., Raghavan, M., Wu, F., Kleinberg, J., and Weinberger, K.~Q. (2017).
\newblock On fairness and calibration.
\newblock {\em Advances in neural information processing systems}, 30.

\bibitem[Rouzot et~al., 2023]{Rouzot2023Learning}
Rouzot, J., Ferry, J., and Huguet, M.-J. (2023).
\newblock Learning optimal fair scoring systems for multi-class classification.
\newblock {\em ICTAI 2022 - The 34\textsuperscript{th} IEEE International
  Conference on Tools with Artificial Intelligence, Oct 2022, Virtual, United
  States}.

\bibitem[Silvia et~al., 2020]{ChiappaRST2020AAAI}
Silvia, C., Ray, J., Tom, S., Aldo, P., Heinrich, J., and John, A. (2020).
\newblock A general approach to fairness with optimal transport.
\newblock In {\em Proceedings of the AAAI Conference on Artificial
  Intelligence}, volume~34, pages 3633--3640.

\bibitem[Tang and Zhang, 2022]{TangZ2022Causality}
Tang, Z. and Zhang, K. (2022).
\newblock Attainability and optimality: The equalized odds fairness revisited.
\newblock In {\em Conference on Causal Learning and Reasoning}, pages 754--786.
  PMLR.

\bibitem[Taturyan et~al., 2024]{TaturyanCH2024NeurIPS}
Taturyan, G., Chzhen, E., and Hebiri, M. (2024).
\newblock Regression under demographic parity constraints via unlabeled
  post-processing.
\newblock {\em Advances in Neural Information Processing Systems},
  37:117917--117953.

\bibitem[Verma and Rubin, 2018]{VermaR2018IWSF}
Verma, S. and Rubin, J. (2018).
\newblock Fairness definitions explained.
\newblock In {\em Proceedings of the International Workshop on Software
  Fairness}, FairWare '18, page 1–7, New York, NY, USA. Association for
  Computing Machinery.

\bibitem[Wang et~al., 2023]{WangHGC2023NeurIPS}
Wang, H., He, L., Gao, R., and Calmon, F. (2023).
\newblock Aleatoric and epistemic discrimination: Fundamental limits of
  fairness interventions.
\newblock {\em Advances in Neural Information Processing Systems},
  36:27040--27062.

\bibitem[Weber et~al., 2022]{weber2022enforcing}
Weber, A., Metevier, B., Brun, Y., Thomas, P.~S., and da~Silva, B.~C. (2022).
\newblock Enforcing delayed-impact fairness guarantees.
\newblock {\em arXiv preprint arXiv:2208.11744}.

\bibitem[Woodworth et~al., 2017]{WoodworthGOS2017COLT}
Woodworth, B., Gunasekar, S., Ohannessian, M.~I., and Srebro, N. (2017).
\newblock Learning non-discriminatory predictors.
\newblock In {\em Conference on learning theory}, pages 1920--1953. PMLR.

\bibitem[Yadav et~al., 2024]{YadavRBC2024NeurIPS}
Yadav, C., Roy~Chowdhury, A., Boneh, D., and Chaudhuri, K. (2024).
\newblock {F}air{P}roof : Confidential and certifiable fairness for neural
  networks.
\newblock In Salakhutdinov, R., Kolter, Z., Heller, K., Weller, A., Oliver, N.,
  Scarlett, J., and Berkenkamp, F., editors, {\em Proceedings of the 41st
  International Conference on Machine Learning}, volume 235 of {\em Proceedings
  of Machine Learning Research}, pages 55682--55705. PMLR.

\bibitem[Zafar et~al., 2019]{ZafarVGG2019JMLR}
Zafar, M.~B., Valera, I., Gomez-Rodriguez, M., and Gummadi, K.~P. (2019).
\newblock Fairness constraints: A flexible approach for fair classification.
\newblock {\em Journal of Machine Learning Research}, 20(75):1--42.

\bibitem[Zoppelletto, 2024]{financialRiskData}
Zoppelletto, L. (2024).
\newblock {Financial Risk for Loan Approval}.
\newblock Kaggle.

\end{thebibliography}

\clearpage
\appendix
\thispagestyle{empty}
\crefalias{section}{appendix}
\crefalias{subsection}{appendix}
\pagenumbering{arabic}
\renewcommand*{\thepage}{A\arabic{page}}
\onecolumn
\conferencetitle{{\mytitle}: \\
  Supplementary Materials}

\section{\uppercase{Proof and Discussion of Theorem}~\ref{thm:main}}
\label{sec:proof-disc-main}

\main*
\begin{proof}[Proof of \cref{thm:main}]
  The constraint ensures that $\tilde{Y}$ is $\eps$-fair, while maximizing the accuracy of $\tilde Y$.
  It remains to show that the optimization problem \cref{eq:main:1} is a linear program.

  First, we argue that $\ell(Y, \tilde Y)$ is an affine function of $p_{\tilde Y|\hat Y Z}$, as
  \begin{align}
    \ell(Y, \tilde Y) & = P\{Y \neq \tilde Y\} = 1 - P\{Y = \tilde Y\}                                                             \\
                      & = 1 - \sum_{y, \hat y, z} p_{\hat YYZ}(\hat y,y,z) p_{\tilde Y|\hat Y Z}(y|\hat y, z) , \label{eq:proof:1}
  \end{align}
  where we used \cref{eq:markov_chain}.
  The implicit constraints, ensuring that $p_{\tilde Y|\hat Y Z}$ is a valid probability mass function can be written as linear inequalities.

  Finally, we can complete the proof by showing that $\mathbb E[\happy(\tilde{Y}, X, Y, Z)|Z=z]$ can be written as a linear function
  \begin{align}
    \mathbb E[\happy(\tilde{Y}, X, Y, Z)|Z=z] = \sum_{\tilde{y}, \hat y} \ehappy(\tilde{y}, \hat y, z) p(\tilde{y} | \hat y, z) , \label{eq:proof:2}
  \end{align}
  where the coefficients are given by
  $
    \ehappy(\tilde{y}, \hat y, z) = p_{\hat Y|Z}(\hat y|z) \,\mathbb E\big[\happy(\tilde y, X, Y, z) \big| \hat Y = \hat y, Z = z\big]
  $, which yields
  \begin{align}
    \sum_{\tilde{y}, \hat y} \ehappy(\tilde{y}, \hat y, z) p(\tilde{y} | \hat y, z) & = \sum_{\tilde{y}, \hat y} p_{\hat Y|Z}(\hat y|z) \,\mathbb E\big[\happy(\tilde y, X, Y, z) \big| \hat Y = \hat y, Z = z\big]  p_{\tilde Y|\hat Y Z}(\tilde{y} | \hat y, z) \\
                                                                                    & = \sum_{\tilde{y}, \hat y} \,\mathbb E\big[\happy(\tilde y, X, Y, z) \big| \hat Y = \hat y, Z = z\big]  p_{\tilde Y \hat Y | Z}(\tilde{y}, \hat y | z)                      \\
                                                                                    & = \mathbb E \Big[ \mathbb E\big[\happy(\tilde Y, X, Y, z) \big| \hat Y, Z = z\big] \Big| Z = z\Big]                                                                         \\
                                                                                    & = \mathbb E\big[\happy(\tilde Y, X, Y, Z) \big| Z = z\big] .
  \end{align}
\end{proof}

\subsection{Empirical Approximation}
\label{sec:empir-appr}

When replacing expectation with empirical expectation in \cref{thm:main}, an accurate result can be obtained if the loss function $\ell(Y, \tilde Y)$ as well as $\ehappy(\tilde{y}, \hat y, z)$ can be well approximated.

We use empirical approximations of $p_{\hat YYZ}(\hat y,y,z)$ as well as $\ehappy(\tilde{y}, \hat y, z)$ defined as
\begin{align}
  \hat p_{\hat YYZ}(\hat y,y,z)     & := \frac{1}{|\mathcal D|} \sum_{(x', y', p_{\hat Y}, z') \in \mathcal D}  p_{\hat Y}(\hat y) \ind_{z'}(z) \ind_{y'}(y) , \label{eq:empirical:1}                               \\
  \hat\ehappy(\tilde{y}, \hat y, z) & := \frac{1}{\hat p_{Z}(z) \, |\mathcal D|} \sum_{(x, y, p_{\hat Y}, z') \in \mathcal D}  \ind_{z'}(z) p_{\hat Y}(\hat y)  \happy(\tilde{y}, x, y, z) , \label{eq:empirical:2}
\end{align}
where $\mathcal D$ is a validation dataset and $\ind$ denotes the indicator function. If the approximations are tight, a bound on the solution of \cref{eq:main:1} can readily be obtained.

\begin{lemma}
  \label{lem:approx1}
  Let $L(\eps)$ be the solution of \cref{eq:main:1} and $\hat L(\eps)$ the solution when substituting with \cref{eq:empirical:1,eq:empirical:2}. Assume that $|\hat p_{\hat YYZ}(\hat y,y,z)] - p_{\hat YYZ}(\hat y,y,z)| \le \delta$ and $|\hat\ehappy(\tilde{y}, \hat y, z)] - \ehappy(\tilde{y}, \hat y, z)| \le \delta$ for all $\tilde{y}, \hat y, y, z$.
  Then, $L(\eps) \le \hat L(\eps - 2\delta) + \delta$.
\end{lemma}
\begin{proof}
  Let $p_{\tilde{Y}|\hat YZ}$ achieve $\hat L(\eps - 2\delta)$. By the assumption, substituting $\hat p$ and $\hat\ehappy$ in \cref{eq:proof:1,eq:proof:2}, results in an error of at most $\delta$, thus yielding $\ell(Y, \tilde Y) \le \hat L(\eps - 2\delta) + \delta$ and
  \begin{align}
    \label{eq:4}
    | \mathbb E[\happy(\tilde{Y}, X, Y, Z)|Z=0] - \mathbb E[\happy(\tilde{Y}, X, Y, Z)|Z=1] | \le \eps .
  \end{align}
\end{proof}

Noting that $\mathbb E[\hat p_{\hat YYZ}(\hat y,y,z)] = p_{\hat YYZ}(\hat y,y,z)$ and $\mathbb E[\hat\ehappy(\tilde{y}, \hat y, z)] = \ehappy(\tilde{y}, \hat y, z)$, there are a total of $2(1+n)|\mathcal Y|^2$ random variables, which are required to be within $\delta$ of their respective expected value for the conditions of \cref{lem:approx1} to be satisfied. Note in particular, that this only depends on the size of the label space, not on the feature spaces. For known $\ehappy$, a concentration result can be used to obtain bounds for the necessary size of the validation set to guarantee adequate approximation with high probability:

\begin{lemma}
  \label{lem:approx2}
  Let $\gamma, \delta > 0$ and assume $A \le \happy \le B$ with $1 \le C := B-A$. Furthermore, let the validation set $\mathcal D$ contain at least $D$ samples from each group.
  Then, if
  \begin{equation}
    \label{eq:boundD}
    D \ge \frac{C^2}{2\delta^2} \log \frac{4(n+1)|\mathcal Y|^2}{\gamma} ,
  \end{equation}
  with probability at least $1-\gamma$, we have that the assumptions of \cref{lem:approx1} are satisfied, i.e., $|\hat p_{\hat YYZ}(\hat y,y,z)] - p_{\hat YYZ}(\hat y,y,z)| \le \delta$ and $|\hat\ehappy(\tilde{y}, \hat y, z)] - \ehappy(\tilde{y}, \hat y, z)| \le \delta$ for all $\tilde{y}, \hat y, y, z$.
\end{lemma}
\begin{proof}
  For each $\hat y,y,z \in \mathcal Y^2 \times \{0,1\}$ let $\mathcal E_{\hat y,y,z}^0$ be the event that $|\hat p_{\hat YYZ}(\hat y,y,z)] - p_{\hat YYZ}(\hat y,y,z)| \ge \delta$. Similarly, let $\mathcal E_{\tilde{y}, \hat y, z}^i$ be the event that $|\hat\ehappy_i(\tilde{y}, \hat y, z)] - \ehappy_i(\tilde{y}, \hat y, z)| \le \delta$ for all $\tilde{y}, \hat y, z,i \in \mathcal Y^2 \times \{0,1\} \times \{1,\dots,n\}$.
  Using Hoeffding's inequality we obtain
  \begin{align*}
    P\{\mathcal E_{\tilde{y}, \hat y, z}^i\} & \le 2\exp\bigg( - \frac{2\delta^2 D}{ C^2 } \bigg) \qquad i \in \{1,\dots,n\}           \\
    P\{\mathcal E_{\hat{y}, y, z}^0\}        & \le 2\exp\big( - 4\delta^2 D \big) \le 2\exp\bigg( - \frac{2\delta^2 D}{ C^2 } \bigg) .
  \end{align*}
  Thus, bounding the probability of the union,
  \begin{align*}
    P\Bigg\{\bigcup_{y \in\mathcal Y,  y'\in\mathcal Y, z \in\mathcal Z, i \in\{0,\dots,n\}} \mathcal E_{y, y', z}^i \Bigg\}
     & \le  2(n+1)|\mathcal Y|^2 2 \exp\bigg( - \frac{2\delta^2 D}{ C^2 } \bigg) \\
     & =  4(n+1)|\mathcal Y|^2 \exp\bigg( - \frac{2\delta^2 D}{ C^2 } \bigg)
  \end{align*}
  and the desired bounds hold with probability at least $1-\gamma$ if \cref{eq:boundD} holds.
\end{proof}

Note that the bound \cref{eq:boundD} yields achievable quantities in typical situations. E.g., for any of the criteria in~\cref{sec:recov-other-crit} and a binary classification problem, we can take $C = 1$ and $n = |\mathcal Y| = 2$. If we require an accuracy of $\delta = 0.02$ with a probability of at least $1-\gamma = 0.99$, the bound \cref{eq:boundD} requires $D \ge \SI{10596}{}$, which is satisfied in the experiments in~\cref{sec:caseStudies}.

\section{\uppercase{Case Study Details}}
\label{sec:details}

\subsection{Case Study with Synthetic Data}
\label{sec:details:synthetic}

\begin{figure}[!htb]
  \centering
  \begin{subfigure}{0.49\linewidth}
    \begin{tikzpicture}
  \begin{axis}[
    grid,
    legend entries={\textbf{Ours}, Equalized Odds, Statistical Parity, Overall Accuracy},
    legend style={
      at={(.05,.05)},
      anchor=south west,
      fill=white,
      font=\small,
      legend image post style={line width=1.5pt}
    },
    x label style={at={(axis description cs:0.0,0.0)},anchor=north},
    ylabel={$\eps(A)$},
    xlabel={$A$},
    ]
    \tikzset{every plot/.append style={mark=none, line width = 1.5pt}}
    \addplot+ table [x=acc, y=0, col sep=comma] {./plots/synthetic/acc_vs_happiness_delta_test.csv};
    \addplot+ table [x=acc, y=1, col sep=comma] {./plots/synthetic/acc_vs_happiness_delta_test.csv};
    \addplot[color=violet] table [x=acc, y=2, col sep=comma] {./plots/synthetic/acc_vs_happiness_delta_test.csv};
    \addplot[color=olive] table [x=acc, y=3, col sep=comma] {./plots/synthetic/acc_vs_happiness_delta_test.csv};
    \end{axis}
\end{tikzpicture}
    \caption{Test}
    \label{fig:synthetic2:test}
  \end{subfigure}
  \hfill
  \begin{subfigure}{0.49\linewidth}
    \begin{tikzpicture}
  \begin{axis}[
    grid,
    legend entries={\textbf{Ours}, Equalized Odds, Statistical Parity, Overall Accuracy},
    legend style={
      at={(.05,.05)},
      anchor=south west,
      fill=white,
      font=\small,
      legend image post style={line width=1.5pt}
    },
    x label style={at={(axis description cs:0.0,0.0)},anchor=north},
    xlabel={$A$},
    ]
    \tikzset{every plot/.append style={mark=none, line width = 1.5pt}}
    \addplot+ table [x=acc, y=0, col sep=comma] {./plots/synthetic/acc_vs_happiness_delta_val.csv};
    \addplot+ table [x=acc, y=1, col sep=comma] {./plots/synthetic/acc_vs_happiness_delta_val.csv};
    \addplot[color=violet] table [x=acc, y=2, col sep=comma] {./plots/synthetic/acc_vs_happiness_delta_val.csv};
    \addplot[color=olive] table [x=acc, y=3, col sep=comma] {./plots/synthetic/acc_vs_happiness_delta_val.csv};
    \end{axis}
\end{tikzpicture}
    \caption{Validation}
    \label{fig:synthetic2:val}
  \end{subfigure}
  \caption{Experiment on \texttt{Synthetic} dataset.}
  \label{fig:synthetic2}
\end{figure}
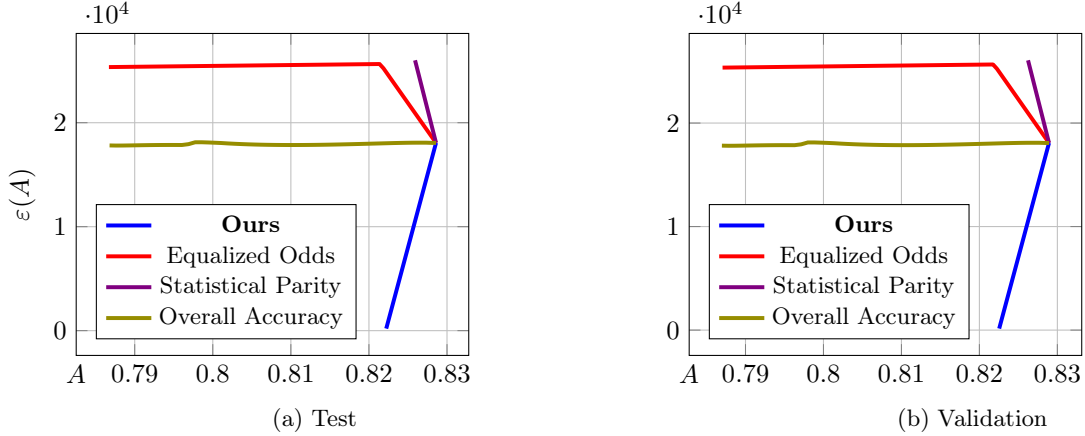

\Cref{fig:synthetic2} shows the results on the problem introduced in~\cref{sec:example}. Values for the test and validation set are very close.

\subsection{Case Study with \texttt{Financial Risk for Loan Approval} Data}
\label{sec:financialRiskDetails}
In this section we report the details for the reproducibility of the experiment in \cref{sec:example_loan_risk}
where:
\begin{itemize}
  \item The step function $\rho(X_{\texttt{credit\_score}})$ is defined as:
        \begin{align*}
          \rho(X_{\texttt{credit\_score}}) =
          \begin{cases}
            0.04 & \text{if } X_{\texttt{credit\_score}} \geq 750       \\
            0.06 & \text{if } 700 \leq X_{\texttt{credit\_score}} < 750 \\
            0.08 & \text{if } 650 \leq X_{\texttt{credit\_score}} < 700 \\
            0.12 & \text{if } 600 \leq X_{\texttt{credit\_score}} < 650 \\
            0.18 & \text{if } X_{\texttt{credit\_score}} < 600          \\
          \end{cases}
        \end{align*}
  \item The return of interest is defined as \[R(X) = \sum_{j\in \{\texttt{loan\_purpose}, \texttt{education}, \texttt{employment}, \texttt{tenure}\}} \beta(X_j),
        \]
        where \(\beta(x)\) are domain-informed weights, assigned as reported below.
  \item The Bonuses $\beta$ are assigned according to the following criteria:
        \begin{itemize}
          \item Loan Purpose Bonuses
                \begin{enumerate}
                  \item $\beta(\texttt{Home}) = 0.08$,
                  \item $\beta(\texttt{Auto}) = 0.02$,
                  \item $\beta(\texttt{Education}) = 0.12$,
                  \item $\beta(\texttt{Debt Consolidation}) = 0.04$,
                  \item $\beta(\texttt{Other}) = 0.05$.
                \end{enumerate}

          \item Education Level Bonuses
                \begin{enumerate}
                  \item $\beta(\texttt{Master}) = 0.01$,
                  \item $\beta(\texttt{Doctorate}) = 0.02$.
                \end{enumerate}

          \item Employment Status Bonuses
                \begin{enumerate}
                  \item $\beta(\texttt{Employed}) = \beta(\texttt{Self-Employed}) = 0.01$.
                \end{enumerate}
          \item Tenure Bonus
                \begin{enumerate}
                  \item $\beta(n) = 0.01 \cdot \ind\{n > 5 \text{ years}\}$.
                \end{enumerate}
        \end{itemize}
\end{itemize}

\end{document}


%
\runningtitle{I use this title instead because the last one was very long}

%

\onecolumn
\aistatstitle{Instructions for Paper Submissions to AISTATS 2026: \\
Supplementary Materials}

\section{FORMATTING INSTRUCTIONS}

To prepare a supplementary pdf file, we ask the authors to use \texttt{aistats2026.sty} as a style file and to follow the same formatting instructions as in the main paper.
The only difference is that the supplementary material must be in a \emph{single-column} format.
You can use \texttt{supplement.tex} in our starter pack as a starting point, or append the supplementary content to the main paper and split the final PDF into two separate files.

Note that reviewers are under no obligation to examine your supplementary material.

\section{MISSING PROOFS}

The supplementary materials may contain detailed proofs of the results that are missing in the main paper.

\subsection{Proof of Lemma 3}

\textit{In this section, we present the detailed proof of Lemma 3 and then [ ... ]}

\section{ADDITIONAL EXPERIMENTS}

If you have additional experimental results, you may include them in the supplementary materials.

\subsection{Effect of the Regularization Parameter}

\textit{Our algorithm depends on the regularization parameter $\lambda$. Figure 1 below illustrates the effect of this parameter on the performance of our algorithm. As we can see, [ ... ]}

\vfill